\documentclass{egpubl}
\usepackage{eurovis-after}

\SpecialIssueSubmission    
 \electronicVersion 

\ifpdf \usepackage[pdftex]{graphicx} \pdfcompresslevel=9
\else \usepackage[dvips]{graphicx} \fi

\PrintedOrElectronic

\usepackage{t1enc,dfadobe}

\usepackage{egweblnk}
\usepackage{cite}

\title[Measuring Visual Complexity of Cluster-Based Visualizations]%
      {Measuring Visual Complexity of Cluster-Based Visualizations}

\author[B. Duffy et al.]{B. Duffy$^1$, A. Dasgupta$^2$, R. Kosara$^2$, S. Walton$^1$, and M. Chen$^1$
         \\
         $^1$ University of Oxford, UK\\
         $^2$ University of North Carolina, Charlotte, USA
}

\graphicspath{{figures/}}

\usepackage{mathptmx}
\usepackage{amsmath}
\usepackage{subfig}
\usepackage{graphicx}
\usepackage{units}
\usepackage{times}
\usepackage{color}
\usepackage{float}
\usepackage{algpseudocode}
\usepackage{algorithm}

\floatname{algorithm}{Procedure}

\newcommand{\etal}{\emph{et al.}~}
\newcommand{\figref}[1]{Figure~\ref{fig:#1}}

\newcommand{\secref}[1]{Section \ref{sec:#1}}

\newcommand{\eqntworef}[2]{Equations \ref{eqn:#1} and \ref{eqn:#2}}
\newcommand{\eqnthreeref}[3]{Equations \ref{eqn:#1}, \ref{eqn:#2} and \ref{eqn:#3}}

\newcommand{\procref}[1]{Procedure \ref{proc:#1}}

\begin{document}

\maketitle

\begin{abstract}
Handling visual complexity is a challenging problem in visualization owing to the subjectiveness of its
definition and the difficulty in devising generalizable quantitative metrics. In this paper we address this
challenge by measuring the visual complexity of two common forms of cluster-based visualizations: scatter
plots and parallel coordinatess. We conceptualize visual complexity as a form of visual uncertainty,
which is a measure of the degree of difficulty for humans to interpret a visual representation correctly.
We propose an algorithm for estimating visual complexity for the aforementioned visualizations using Allen's
interval algebra. We first establish a set of primitive 2-cluster cases in scatter plots and another set for
parallel coordinatess based on symmetric isomorphism. We confirm that both are the minimal sets and
verify the correctness of their members computationally. We score the uncertainty of each primitive case
based on its topological properties, including the existence of overlapping regions, splitting regions and
meeting points or edges. We compare a few optional scoring schemes against a set of subjective scores by
humans, and identify the one that is the most consistent with the subjective scores. Finally, we extend the
2-cluster measure to k-cluster measure as a general purpose estimator of visual complexity for these two
forms of cluster-based visualization.
\end{abstract}

\section{Introduction}
Visual complexity is a pervasive problem in different domains such as graphical user interfaces, web information, visualizations, etc.
While the correlation between visual complexity and cognitive load \cite{Harper:2009:TDV:} has been established, it is widely acknowledged 
that one of the main challenges is to provide an objective definition such that it bridges system-level behavior with user perception \cite{schnur2010comparison}.
The subjectiveness of this notion makes it difficult to develop reliable metrics for measuring visual complexity.

The focus of this paper is to measure \emph{visual complexity} in cluster visualization.
We examine two forms of such visualization, namely scatter plots and parallel coordinatess \cite{Inselberg1990}.
Here we define visual complexity as a form of \emph{visual uncertainty}\cite{dasgupta:2012}.
It measures visual components, such as overlapped points, lines and shapes, missing objects, and split or disconnected shapes, that may lead to confusion in viewing the visualization.
Our contributions are:
\begin{itemize}
\item We propose a novel application of Allen's interval algebra for formulating a metric for measuring visual complexity.
\item We show that the $13 \times 13$ topological cases in 2D can be reduced to 24 primitive cases for scatter plots and 35 primitive cases for parallel coordinates.
\item We define two metrics for estimating visual complexity in scatter plots and parallel coordinates respectively and we make use of look-up tables in a manner similar to the marching cubes algorithm \cite{lorensen1987}.
\item We compare the scores of the two metrics with a set of subjective scores by humans, and confirm the two metrics are effective. 
\end{itemize}

\section{Related Work}
\label{sec:related_work}
In this section we discuss the relevant literature on clustered parallel coordinates
and scatter plots, and on concepts of and metrics for visual complexity.

\subsection{Clustered Scatter Plots and Parallel Coordinates}
Traditional clustering techniques in visualization are of two main types: analytical clustering and visual
clustering. Analytical clustering aims to maximize within cluster information by using data-space properties
\cite{ Fua:Vis:1999, Novotny2006}. Two-dimensional clusters that tend to overlap between axes
\cite{Andrienko2004} add to the visual complexity, but computational approaches to quantify that has been
absent in the literature. Visual clustering in parallel coordinates aims to reduce clutter; some examples
include geometrically deforming and grouping poly-lines to overcome edge clutter \cite{Zhou2008} and use of
high-precision textures for reducing the effect of over-plotting \cite{Johansson}. Privacy-preserving
clustering \cite{Dasgupta:InfoVis:2011} encompasses these two categories, the goal here being controlling
the within-cluster information to prevent disclosure. Quantifying complexity in terms of uncertainty
measures have been found to be useful in quantifying the utility of privacy-preserving visualizations~[withheld].

\subsection{Visual Complexity Measures}
Rosenholtz et al.~\cite{rosenholtz2007} describe a number of methods for measuring visual clutter and complexity based on the
ideas of feature congestion and reaction time. They highlight the current state of the art for measuring
visual complexity falls into two categories. Simplistic measures of visual complexity based on counting
geometric primitives such as lines and triangles, and complex measures based on computer vision techniques.
However, these methods have a number of drawbacks. The simplistic methods are generally used for
visualization displays in two and three dimensions, making them dependant on the input data. In addition
there is only a weak correlation between the number of primitives in the display and the complexity of the
visual output \cite{carr2006,scheidegger2008,khoury2010,duffy2012}. The complex methods are computationally
intense and not appropriate for visualization displays when access to raw geometric data is available. In
general there is a lack of tools for measuring visual complexity in visualization applications that
quantifies overlap and occlusion. Simplistic methods such as, counting geometric primitives, i.e., 
vertices, lines etc. have been used in visualization applications. This method has been applied recently by
Carr \etal \cite{carr2006}, Scheidegger \etal \cite{scheidegger2008} and Duffy \etal \cite{duffy2012} for
measuring the complexity of isosurfaces through triangle counts and cell intersections. Khoury \etal
\cite{khoury2010} use fractal box dimensions to measure isosurface complexity. More complex computer vision
methods are the alternative as illustrated by Rosenholtz \etal \cite{rosenholtz2007}.

\subsection{Related Approaches: Clutter and Visual Quality}
Clutter reduction techniques are important in the context of information visualization as they visual
quality preserving rendering. Ellis and Dix have outlined in their taxonomy \cite{ellis} how the different
clutter reduction approaches fit in a common framework. There is a lack of agreeable definition of clutter
\cite{Ellis2006} and visual quality \cite{Bertini2006}. While there have been approaches to define clutter
in terms of outliers \cite{Peng2004}, other researcher have defined clutter in terms of overlapping visual
objects \cite{Artero2004, Dasgupta:InfoVis:2010}. Similarly with visual quality, while quality metrics have
been proposed to improve the perceptual aspect of visualizations, similar metrics have been suggested for
pattern identification. We believe a decomposition of visualization in terms of its smallest components,
that is, the visual structures will enable us to standardize metrics across different visual representations.
Moreover, quantification of complexity will also enable more concrete optimization processes that can
minimize clutter on screen.

\section{Allen's Interval Algebra}
\label{sec:1D}

\begin{figure}[!t]
\centering
\includegraphics[width=1.0\linewidth]{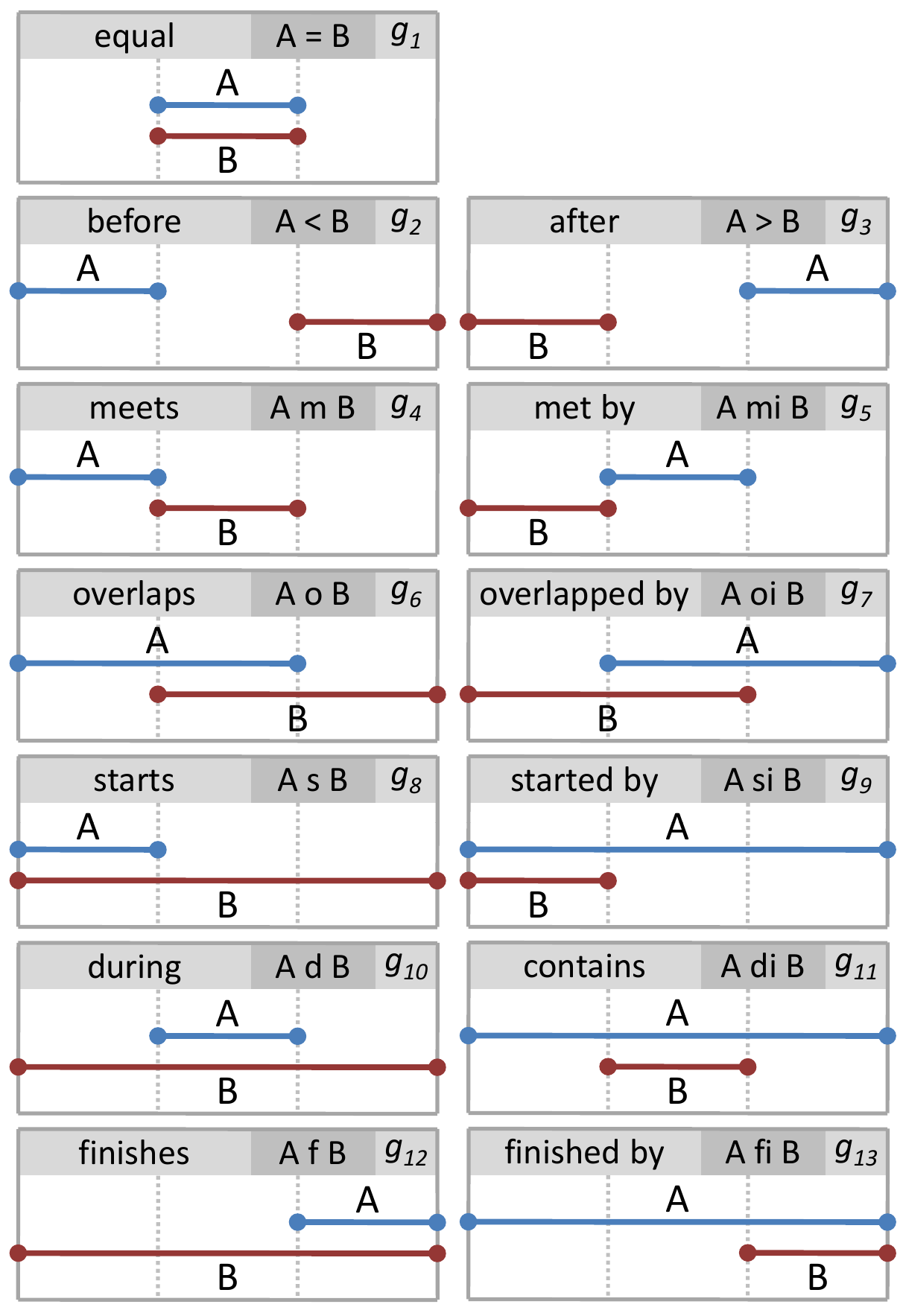}
\caption{Allen algebra intervals in 1D. Shown are the 13 operators in the algebra.}
\label{fig:fig1}
\end{figure}

Allen developed an interval algebra in 1983 for reasoning about discrete time intervals \cite{allen1983}.
As shown \figref{fig1}, the algebra defines a set of 13 operators on two interval operands in 1D. It is not
difficult to observe that the operators exhibit some symmetry in relation to the ordering of the two operands.
For the convenience mathematical representation, let us write each operation in a functional form akin to the
Polish prefix notation:
\[
    g_i(A, B), \quad i=1,2,\ldots, 13
\]
\noindent where $g_i$ is an operator (i.e., $g_1$ is $<$, $g_2$ is $>$, etc.), and $A$ and $B$ are the two
interval operands, $[a_1, a_2]$ and $[b_1, b_2]$, such that $a_1 < a_2$ and $b_1 < b_2$. Note that the
function $g_i$ can be regarded as a Boolean function that determines whether Allen's $i^{th}$ relationship
between $A$ and $B$ is true or false. The \emph{Operand Ordering Symmetry} can thus be expressed as:
\begin{equation}
	\Psi_{OOS} \bigl ( g_i(A, B) \bigr ) \rightarrow g_j(B, A), \quad 1 \leq i, j, \leq 13
	\label{eqn:oos}
\end{equation}
\noindent where $\Psi_{OOS}$ is the \emph{transformation} of swapping the two operands for a given
$g_i(A, B)$. A symmetric relation holds if $g_j$ exists. There are seven pairs of such symmetry, including
the self-symmetry $g_{1}(A, B) = g_{1}(B, A)$.

Another form of symmetry results from flipping an axis towards the opposite direction. In 1D case, given an
interval $X=[x_1, x_2]$, we denote its mirror on the flipped axis as $X^- = [-x_2, -x_1]$.
Hence, the \emph{Axis Flipping Symmetry} can be expressed as: 
\begin{equation}
	\Psi_{AFS} \bigl ( g_i(A, B) \bigr ) \rightarrow g_j(A^-, B^-), \quad 1 \leq i, j, \leq 13
	\label{eqn:afs}
\end{equation}
\noindent where $\Psi_{AFS}$ is the transformation of flipping the axis. There are seven pairs of such
symmetry, including $g_8(A, B) =g_{12}(A^-, B^-)$ and $g_9(A, B) = g_{13}(A^-, B^-)$.

With these two types of symmetry, we can reduce the 13 cases to 6 primitive cases, which are $g_{1}, g_{2},
g_{4}, g_{8}, g_{10}$ (=, <, m, o, s, d). Each of the other 7 cases can be inferred from a primitive case
using one of the two symmetry relations.

\section{2-Cluster Overlaps}
\label{sec:2D}
\begin{figure}[!t]
\centering
\includegraphics[width=1.0\linewidth]{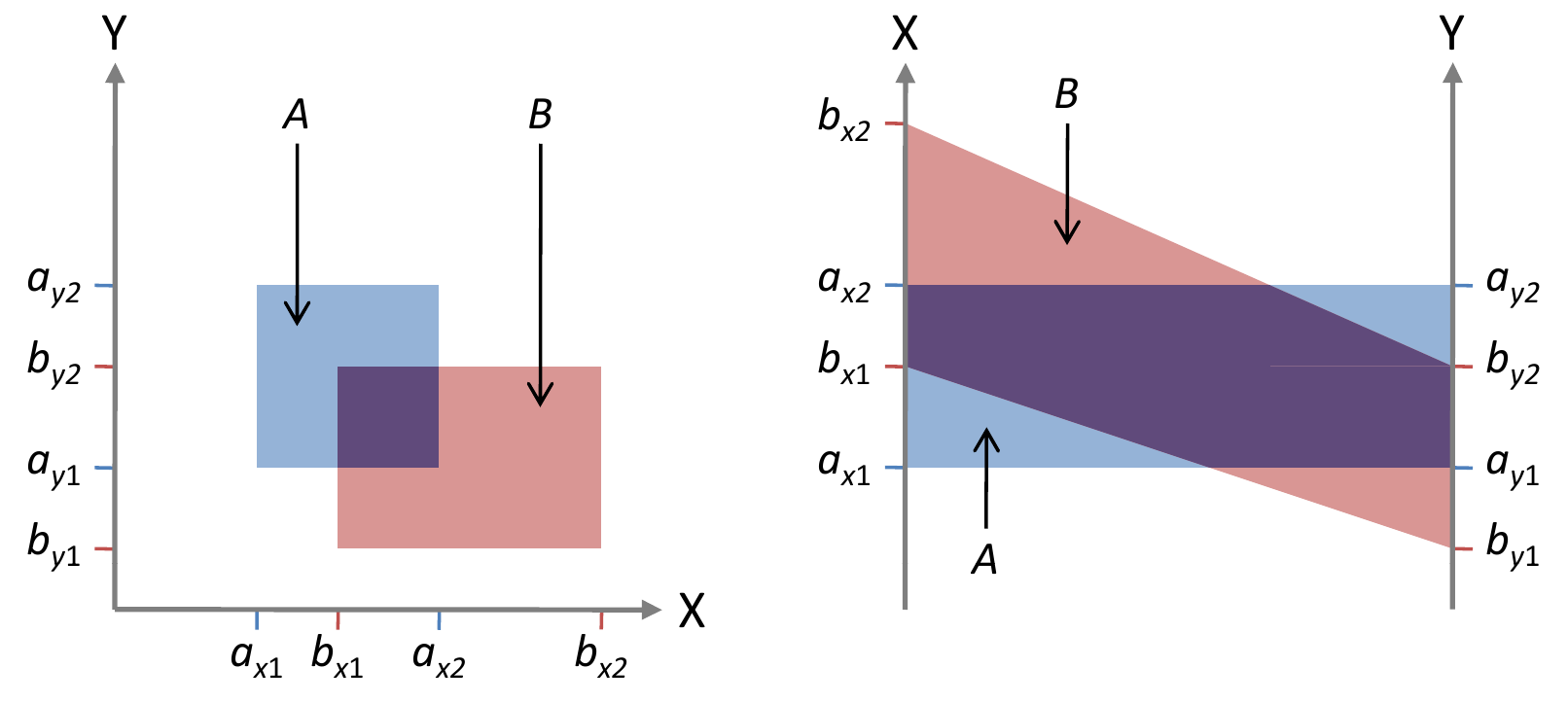}
\caption{Simple case of overlapping clusters in scatter plots and parallel coordinates.}
\label{fig:fig2} 
\end{figure}

Allen's interval algebra can be extended to 2D when examining cases in two common
forms of cluster visualization; namely scatter plots and parallel coordinates. In
previous work Dasgupta and Kosara \cite{Dasgupta:InfoVis:2010} used Allen's algebra
for computing metrics for parallel coordinates. \figref{fig2} shows a simple case of
two overlapping clusters in a scatter plot as well as a parallel coordinates.
The relationship on the $x$-axis is $A o B$ or $g_5(A, B)$, and that on the $y$-axis
is $A oi B$ or $g_6(A, B)$. We can represent this case by the following 2-tuple:
\[
	\bigl [ g_5(A, B), g_6(A, B) \bigr ]
\]
It is not difficult to observe that given an ordered pair of operands, there are $13 \times 13 = 169$
different tuples in 2D.

\subsection{Symmetries in 2D and Primitive Cases}
\begin{figure*}[!t]
	\centering
	\includegraphics[width=1.0\linewidth]{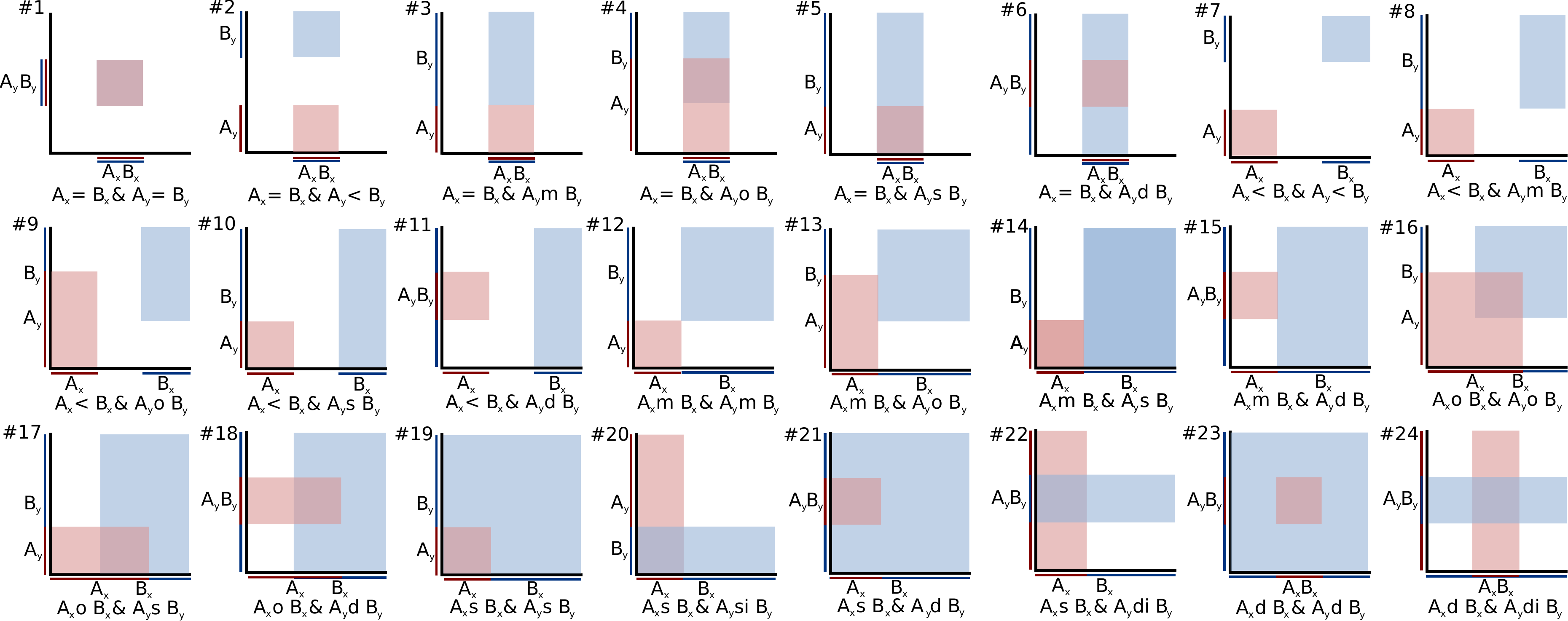}
	\caption{ 169 scatter plot cases can be reduced to a subset of 24 topologically distinct bases cases using
	          4 symmetries. }
	\label{fig:fig3}
\end{figure*}
\begin{figure*}[!t]
\centering
\includegraphics[width=1.0\linewidth]{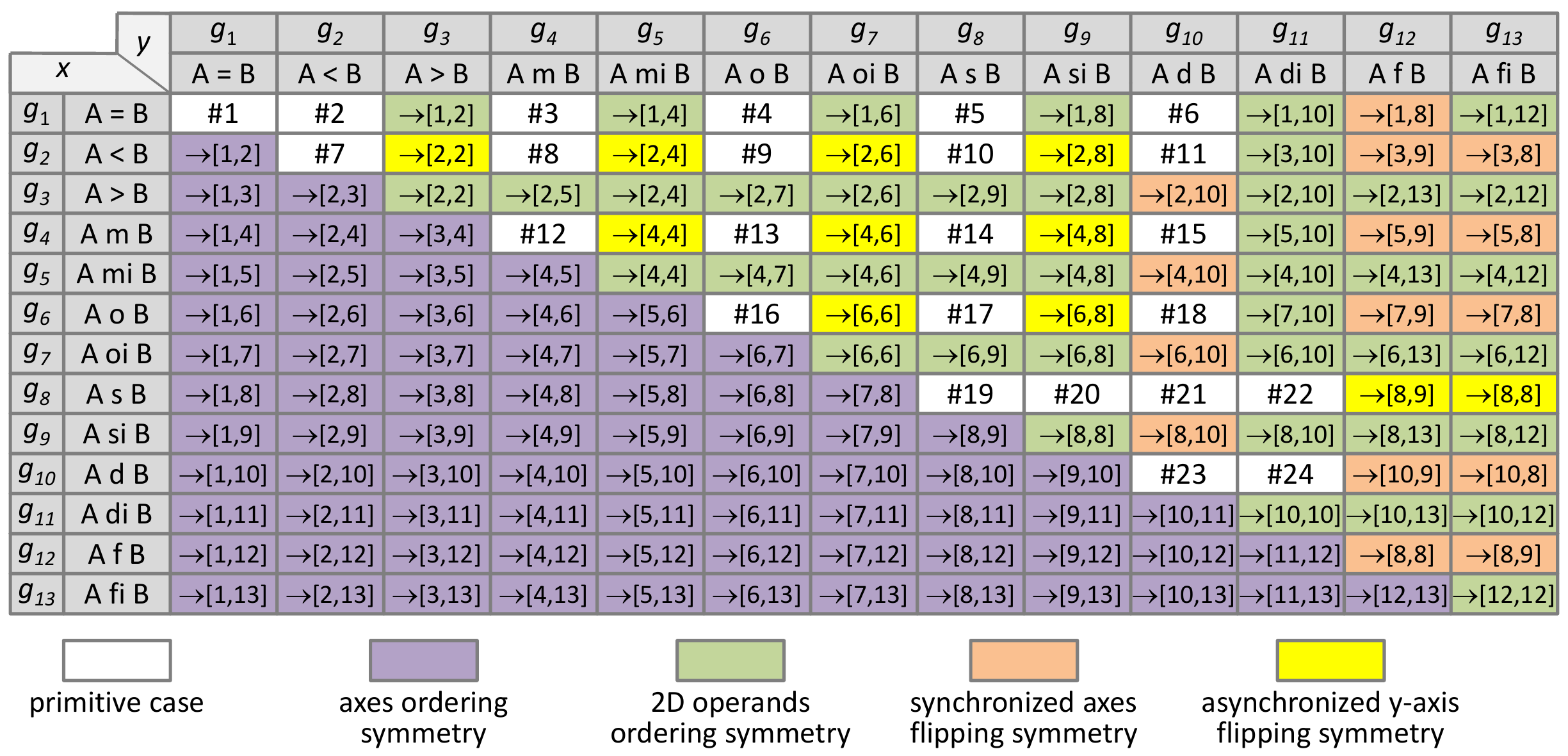}
\caption{The $13 \times 13$ cases of 2D Allen's interval algebra. It shows 24 primitive cases for scatter
plots as numbered in \figref{fig3}, and a transformation path from each of other cases to one of the primitive cases.}
\label{fig:fig4}
\end{figure*}
Using symmetry relationships, we have found that the 169 cases can be reduced to 24 primitive cases in
scatter plots, and 35 primitive cases in parallel coordinates. The symmetry relationships
shared by both types of plots are:

\noindent \textbf{2D Operand Ordering Symmetry} --- This is a direct extrapolation from the same type of
symmetry in 1D. Let $A$ and $B$ be two clusters, their ranges on the $x$-axis are $A_x$ and $B_x$, and those
on the $y$-axis are $A_y$ and $B_y$ respectively. We can express this symmetry in 2D using a transformation
$\Psi_{2d-OOS}$ as:
\begin{equation}
    \begin{split}
	& \Psi_{2d-OOS} \bigl ( \bigl [ g_i(A_x, B_x), g_s(A_y, B_y) \bigr ] \bigr )  \\
	& \rightarrow \bigl [ \Psi_{OOS} \bigl ( g_i(A_x, B_x) \bigr ), \Psi_{OOS} \bigl ( g_s(A_y, B_y) \bigr ) \bigr ]\\
	& \rightarrow \bigl [ g_j(B_x, A_x), g_t(B_y, A_y) \bigr ]
    \end{split}
 	\label{eqn:2doos}
\end{equation}
\noindent where $1 \leq i, j, s, t \leq 13$, and $\Psi_{OOS}$ is the corresponding 1D transformation as
$B$ before and after the symmetric transformation, and similarly $g_s$ and $g_t$ for the $y$-axis.

\noindent \textbf{Synchronized Axes Flipping Symmetry} --- We can also extrapolate the \emph{Axis Flipping
Symmetry} to 2D by flipping both axes simultaneously towards the opposite direction. We can express this
symmetry as:
\begin{equation}
    \begin{split}
	& \Psi_{SAFS} \bigl ( \bigl [ g_i(A_x, B_x), g_s(A_y, B_y) \bigr ] \bigr ) \\
	& \rightarrow \bigl [ \Psi_{AFS} \bigl ( g_i(A_x, B_x) \bigr ), \Psi_{AFS} \bigl ( g_s(A_y, B_y) \bigr ) \bigr ]\\
	& \rightarrow \bigl [ g_j(A_x^-, B_x^-), g_t(A_y^-, B_y^-) \bigr ] 
    \end{split}
  \label{eqn:safs}
\end{equation}
\noindent where $i, j, s, t, g_i, g_j, g_s, g_t$ are defined as previously with $\Psi_{2d-OOS}$.

\noindent \textbf{Axes Ordering Symmetry} --- This is a new form of symmetry in 2D, which encodes the
symmetric transformation, with which the orders of the two axes, $X$ and $Y$ are swapped in the visualization.
\begin{equation}
	\Psi_{AOS} \bigl ( \bigl [ g_i(A_x, B_x), g_s(A_y, B_y) \bigr ] \bigr ) \rightarrow
	\bigl [ g_j(A_y, B_y), g_t(A_x, B_x) \bigr ]
	\label{eqn:aos}
\end{equation}
\noindent where $i, j, s, t, g_i, g_j, g_s, g_t$ are defined as previously, except that $g_j$ now applies
to the intervals on the $y$-axis, while $g_t$ on the $x$-axis.

In addition, there is another type of symmetry that is more meaningful to scatter plots than to parallel
coordinates plots. With scatter plots, if one flips either of the two axes individually, it does not change
the topology or amount of overlapping between the two clusters, and thereby has limited impact on the
perception of the visual complexity. On the contrary, flipping only one axis may cause a change of
overlapping relationship in a parallel coordinates. Given two non-overlapping clusters, they would
become overlapped after one of the two axes is flipped. Hence the following symmetry applies only to scatter
plots.

\begin{figure*}[!t]
\centering
\includegraphics[width=1.0\linewidth]{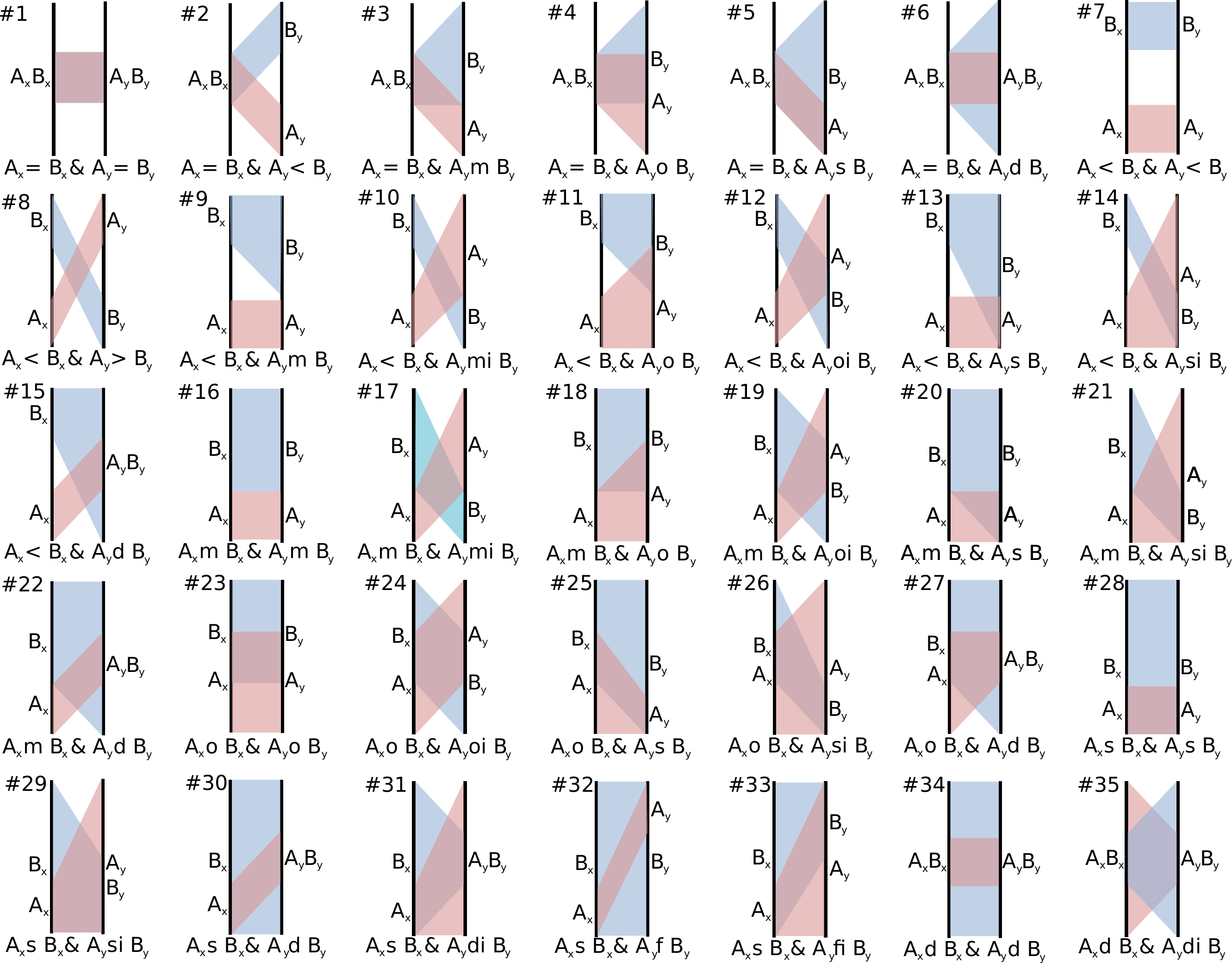}
\caption{169 parallel coordinates cases can be reduced to a subset of 35
topologically distinct base cases using 4 symmetries. }
\label{fig:fig5}
\end{figure*}

\begin{figure*}[!ht]
\centering
\includegraphics[width=1.0\linewidth]{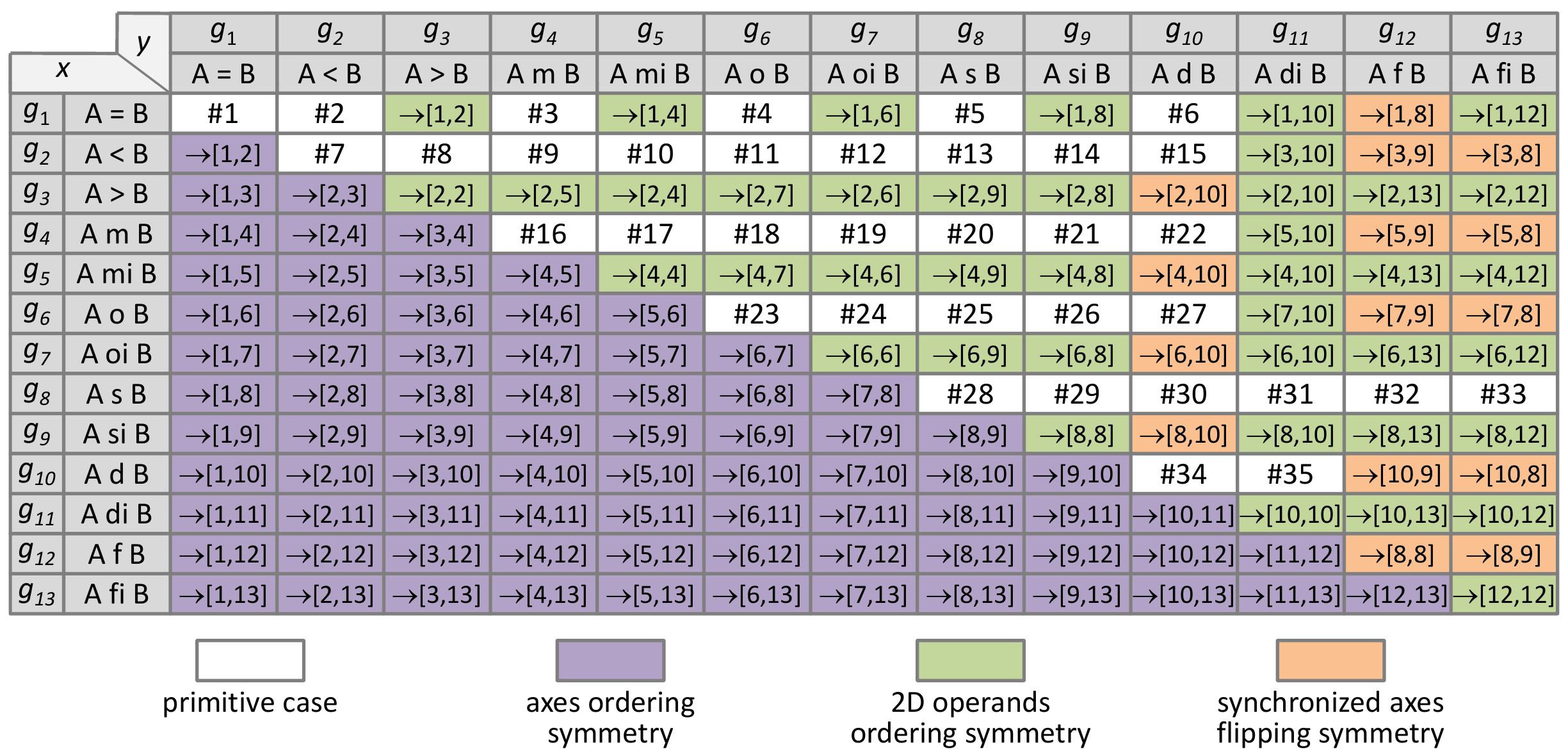}
\caption{ The $13 \times 13$ cases of 2D Allen's interval algebra. It shows 35 primitive cases for parallel
coordinates plots as numbered in \figref{fig5}, and a transformation path from each of other cases to one of the
primitive cases. }
\label{fig:fig6}
\end{figure*}

\noindent \textbf{Asynchronized Axis Flipping Symmetry} --- We use $\Psi_{AxFS}$ to denote the transformation
of flipping the $x$-axis, and $\Psi_{AyFS}$ for that of the $y$-axis. Similar to $\Psi_{SAFS}$, these two
transformations can be expressed as follows:
\begin{equation}
    \begin{split}
	& \Psi_{AXFS} \bigl ( \bigl [ g_i(A_x, B_x), g_s(A_y, B_y) \bigr ] \bigr )            \\ 
	& \rightarrow \bigl [ \Psi_{AFS} \bigl ( g_i(A_x, B_x) \bigr ), g_s(A_y, B_y) \bigr ] \\
	& \rightarrow \bigl [ g_j(A_x^-, B_x^-), g_s(A_y, B_y) \bigr ]
    \end{split}
    \label{eqn:axfs}
\end{equation}

\begin{equation}
    \begin{split}
	& \Psi_{AYFS} \bigl ( \bigl [ g_i(A_x, B_x), g_s(A_y, B_y) \bigr ] \bigr )            \\ 
	& \rightarrow \bigl [ g_i(A_x, B_x), \Psi_{AFS} \bigl ( g_s(A_y, B_y) \bigr ) \bigr ] \\
	& \rightarrow \bigl [ g_j(A_x, B_x), g_t(A_y^-, B_y^-) \bigr ]
    \end{split}
    \label{eqn:ayfs}
\end{equation}
When one of the 169 cases can be transformed to another using any above transformation, they are said to be
\emph{topologically isomorphic}. Since it is relatively trivial to prove that all above-mentioned
transformations are communicative, such a isomorphism is \emph{symmetric}. When a number of cases form an
\emph{isomorphic group}, where each case can be transformed to another through one or more transformations.
For each isomorphic group, we can select one case as the primitive case. 
\figref{fig3} shows 24 primitive cases of Allen's interval algebra in 2D for scatter plots.
\figref{fig4} illustrates some of the symmetric transformations that lead to the formation of these 24
isomorphic groups. \figref{fig5} shows 35 primitives cases for parallel coordinatess, while
\figref{fig6} illustrates the formation of the isomorphic groups.

\subsection{Computational Verification of the Primitive Cases}
We established the isomorphic groups using two different methods.
Firstly we used the matrices in \figref{fig3} and \figref{fig4} as exhaustive lists
of all cases in the two types of plots respective.
We sketched out many cases to identify symmetric transformation from one another.
Secondly, we enumerated all possible symmetric transformations computationally,
providing a verification of the isomorphic groups found manually.
The algorithm for forming each isomorphic group by searching for all possible
symmetric transformations is described below.
\begin{algorithm}
\caption{Exhaustive isomorphic elimination in 1D.}\label{proc:verify1d}
 \algsetblock[Name]{Procedure}{EndProcedure}{16}{3mm}
 \algsetblock[Name]{For}{EndFor}{16}{3mm}
 \algsetblock[Name]{If}{EndIf}{16}{3mm}
	\begin{algorithmic}[1]
	\Procedure{VERIFY1D}{}
		\State $F[1..13] \gets 0$	\Comment{initialize all non-isomorphic}
		\For{each $g_i \in$ Operator Set}
			\If{F[i] = 0}
				\For{each rule $\Psi \in$ Rule Set}
					\State $[h, U, V] \leftarrow \Psi (g_i, A_i, B_i)$ \Comment{transform}
					\For{each $k \in [1..13] \land k \ne i$}
						\If {     
								  $F[k] = 0 \land  EQ([h, U, V], [g_k, A_k, B_k])$
								}
							\State $F[k] \gets i$	\Comment{set isomorphic link}
						\EndIf
					\EndFor
				\EndFor
			\EndIf
		\EndFor
	\EndProcedure
	\end{algorithmic}
\end{algorithm}
The algorithm demonstrates the establishment of isomorphic groups in 1D.
Consider the list of 13 cases, each with an operator $g_i$, as in \figref{fig1}.
\procref{verify1d} exhaustively visits each non-isomorphic case, and applies the rules in the rule set,
$\{ \Psi_{OOS}, \Psi_{AFS}, \Psi_{OOS} \circ \Psi_{AFS}, \}$ based on \eqntworef{oos}{afs},
where $\circ$ denotes the applications of two rules (right first).
If the application of a rule to $g_i(A_i, B_i)$ resulting in $h(U,V)$ that is topologically equitant to another case $g_k$, then $g_k$ is an isomorphic with $g_i$ and $g_k$ is eliminated for further consideration.

\procref{verify2d} shows an algorithm that exhaustively searches isomorphic group in 2D for scatter plots and parallel coordinates.
The rule set for parallel coordinatess are
based on \eqnthreeref{2doos}{safs}{aos}, resulting in
$\{ \Psi_{2d-OOS},\Psi_{SAFS},\Psi_{AOS}$,
$\Psi_{2d-OOS}\circ\Psi_{SAFS}$,
$\Psi_{2d-OOS}\circ\Psi_{AOS}$,
$\Psi_{SAFS}\circ\Psi_{AOS}$,
$\Psi_{2d-OOS}\circ\Psi_{SAFS}\circ\Psi_{AOS}\}$
These rules are also used for scatter plots.
However to achieve full reduction, further rules are required based on \eqntworef{axfs}{ayfs}, including
$\{ \Psi_{AXAFS}$,
$\Psi_{AYAFS}$,
$\Psi_{2d-OOS}\circ\Psi_{AXAFS}$, 
$\Psi_{2d-OOS}\circ\Psi_{AYAFS}$, 
$\Psi_{AXAFS}\circ\Psi_{AOS}$,
$\Psi_{AYAFS}\circ\Psi_{AOS}$,
$\Psi_{2d-OOS}\circ\Psi_{AXAFS}\circ \Psi_{AOS}$,
$\Psi_{2d-OOS}\circ\Psi_{AYAFS}\circ \Psi_{AOS}
\}$
. The rule set does not contain all combinations of the rules because commutative laws apply.
In addition, we have $\Psi_{SAFS} \circ \Psi_{AXFAS} = \Psi_{AYFAS}$ and so on.
\begin{algorithm}
\caption{Exhaustive isomorphic elimination in 2D.}\label{proc:verify2d}
 \algsetblock[Name]{Procedure}{EndProcedure}{16}{3mm}
 \algsetblock[Name]{For}{EndFor}{16}{3mm}
 \algsetblock[Name]{If}{EndIf}{16}{3mm}
	\begin{algorithmic}[1]
	\Procedure{VERIFY2D}{}
		\State $F[1..13][1..13] \gets 0$	\Comment{initialize all non-isomorphic}
		\For{each $g_i \in$ Operator Set}
			\For{each $g_s \in$ Operator Set}
				\If{F[i][s] = 0}
					\For{each rule $\Psi \in$ Rule Set}
						\State $H \gets [h_x, U_x, V_x; h_y, U_y, V_y, ]$
						\State $G \gets [g_{i,x}, A_{i,x}, B_{i,x}, g_{s,y}, A_{i,y}, B_{i,y}]$
						\State $H \gets \Psi (G)$	\Comment{transform}
						\For{each $k,l \in [1..13][1..13] \land k,l \ne i,s$}
							\If {     
								  $F[k][l] = 0 \land  EQ(H,G)$
								}
							\State $F[k][l] \gets <i,s>$	\Comment{set isomorphic link}
						\EndIf
					\EndFor
				\EndFor
			\EndIf
		\EndFor
	\EndProcedure
	\end{algorithmic}
\end{algorithm}
Running \procref{verify2d} confirmed 24 primitive cases for scatter plots in 
\figref{fig3} and 35 primitive cases for parallel coordinates in \figref{fig5}.

\section{Estimating Visual Complexity}
\label{sec:Estimation}
Given a relatively small number of primitive cases in either cluster-based scatter plot
or parallel coordinates, we can consider the notion of \emph{visual complexity}
in a relatively abstract manner by focusing on topological differences between these
cases. In this section, we first propose a scheme for estimating a \emph{complexity
score} for each primitive case. We then compare the scores with a collection of
samples that record how human observers may perceive visual complexity. Finally, we
provide a means for approximating $n$-cluster visual complexity.

\subsection{Estimating 2-cluster Complexity}
\label{sec:2-Cluster}
The purpose of estimating visual complexity is to provide a metric for measuring some
aspects of visual uncertainty as discussed in \cite{dasgupta:2012}. Allen's
interval algebra takes into account both overlapping and ``meeting'' clusters as topological
features. Hence an estimation scheme must encode both features, and it may have the
following principal considerations:
\begin{enumerate}
\item A primitive case should receive the lowest complexity score if it consists of
two clusters that neither overlap nor meet with each other. We make 0 the lowest
complexity score. 
\item A primitive case should receive the highest complexity score if it consists two
clusters that are equal on both axes, i.e., totally coinciding with one another. We
make 1 the highest complexity score.
\item When shape $A$ is not overlapped by shape $B$, $A$ is visually less complex than
when it is crossed over by shape $B$.
\item When shape $A$ has at least one non-overlapping region, $A$ is visually less
complex than when it is totally overlapped by $B$.

\item When shape $A$ is split by shape $B$ into three pieces (1 overlapping and 2
non-overlapping), $A$ is visually more complex than when $A$ is split by $B$ into two
pieces (1 overlapping and 1 non-overlapping).
\item When $A$ and $B$ meet at $k+1$ corners, the case is visually more complex than
when they meet at $k$ corners ($k > 0$).
\item When $A$ and $B$ meet only at a corner, the case is visually less complex than
when they meet along an edge.
\end{enumerate}
\noindent \textbf{Scoring the 24 Primitive Cases of Scatter Plots.}
Given two rectangular shapes $A$ and $B$ representing two clusters in a scatter plot,
we consider an estimation scheme that decomposes a complexity score $U$ into four
components as $U = U_A + U_B + U_{AB} + U_m$.
\begin{itemize}
\item $U_A = 0.0$ if shape $A$ has one continuous non-overlapping region. 
$U_A = 0.1$ if $A$ has two disconnected non-overlapping regions.
$U_A = 0.2$ if $A$ has no non-overlapping region at all.
\item $U_B$ is scored in the same way as $U_A$ by exchanging the relationship between $A$ and $B$.
\item $U_{AB} = 0.0$ if $A$ and $B$ do not overlap, and $U_{AB} = 0.2$ otherwise (i.e., there is one overlapping region).
\item $U_m = 0.1 \times n_e$ where $n_e$ is the number of edges where $A$ and $B$ meet.
$U_m = 0.1$ if $A$ and $B$ do not meet at any edge but at a corner point.
\end{itemize}

\begin{figure}[ht]
\centering
\includegraphics[width=\linewidth]{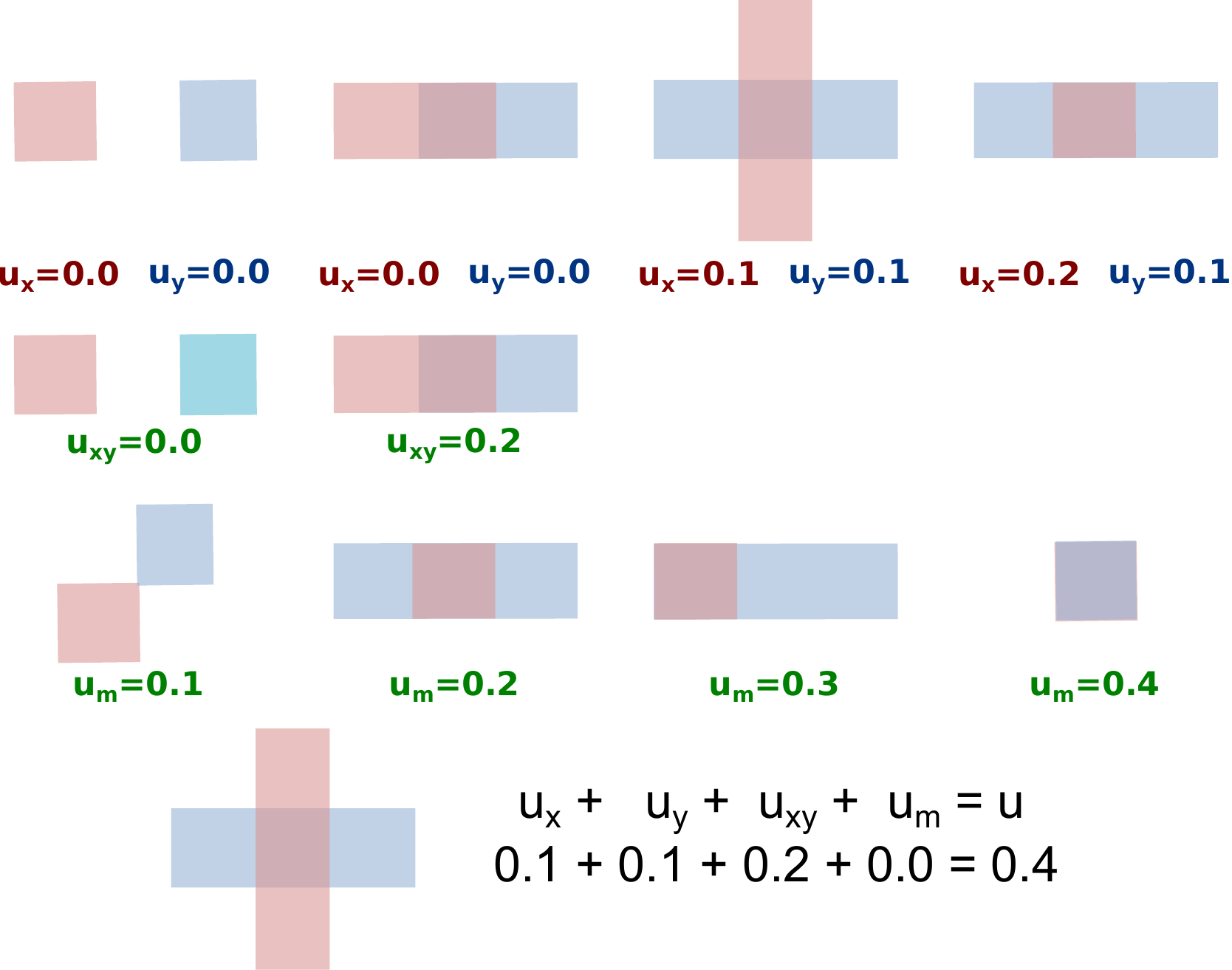}
\caption{Uncertainty approximation for scatter plots.}
\label{fig:fig7}
\end{figure}

\figref{fig7} show some examples that illustrate the scores of $U_A$, $U_B$, $U_{AB}$
and $U_m$ individually. \figref{fig9} lists the scores of $U$ for all $24$ primitive cases of
scatter plots. Note that when $A$ and $B$ coincide completely, $U$ sums up to exactly 1.

\noindent \textbf{Scoring the 35 Primitive Cases of Parallel Coordinates.}
Given two quadrilateral or triangular shapes $A$ and $B$ representing two clusters in
a parallel plot, we consider a similar estimation scheme that decomposes a
complexity score into four components. The first three components $U_A$, $U_B$ and
$U_{AB}$ are computed in the same way as with scatter plots. $U_m$ is computed is a
slightly different way.
\begin{itemize}
\item $U_m = 0.1 \times n_p$ where $n_p$ is the number of corner points where $A$ and $B$ meet.
\end{itemize}

\begin{figure}[ht]
\centering
\includegraphics[width=\linewidth]{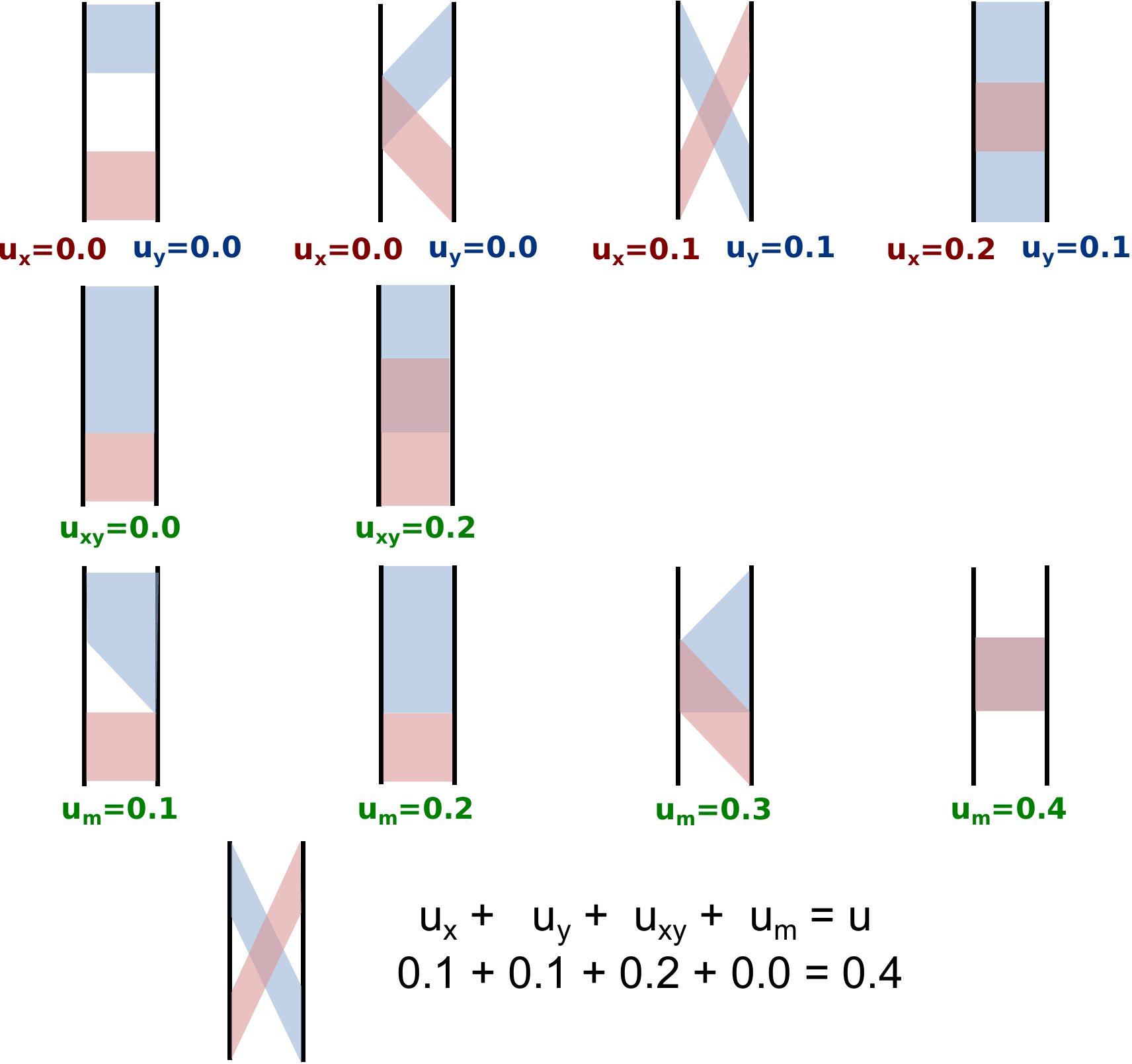}
\caption{Uncertainty approximation for parallel coordinates.}
\label{fig:fig8} 
\end{figure}

\begin{figure*}[ht]
\centering
\includegraphics[trim = 32mm 11mm 30mm 11mm, clip=true, width=\linewidth]{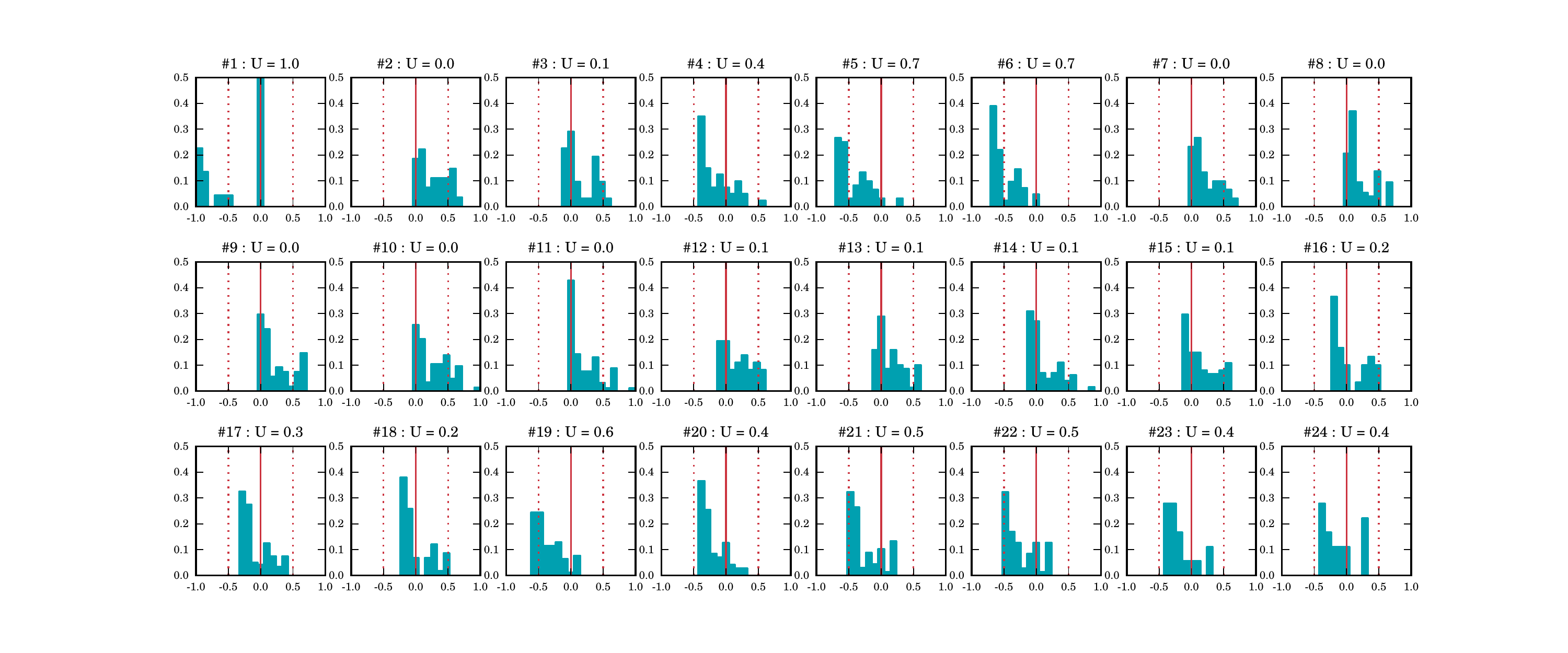}
\caption{Scatter plot case distributions show the amount of agreement of 
measured percieved visual complexity with our scoring system. The scatter score 
is in the title of each plot.}
\label{fig:fig9}
\end{figure*}

\begin{figure*}[ht]
\centering
\includegraphics[trim = 30mm 18mm 28mm 19mm, clip=true, width=\linewidth]{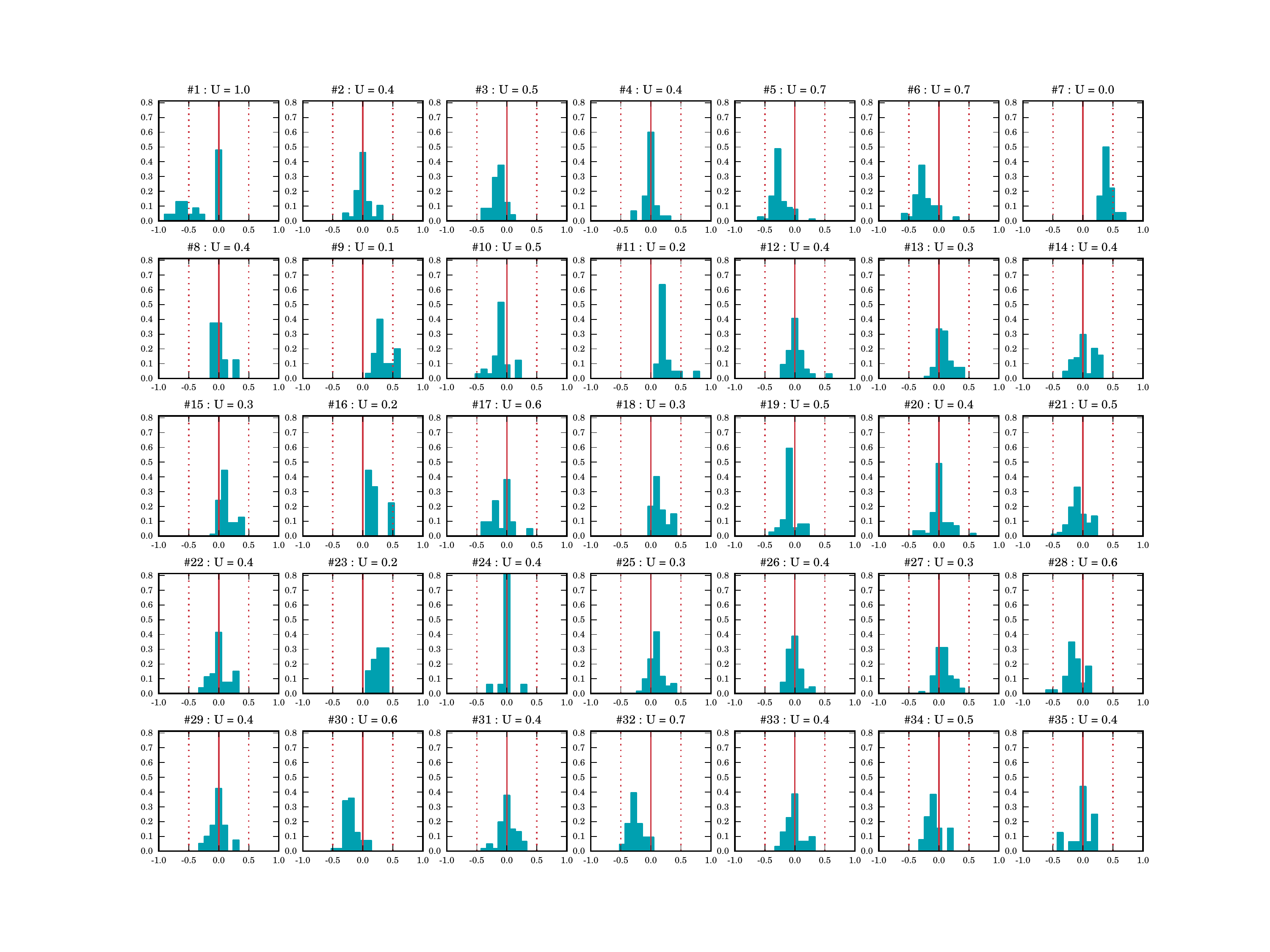}
\caption{Parallel coordinates case distributions show the amount of agreement of 
measured percieved visual complexity with our scoring system. The parallel score 
is in the title of each plot.}
\label{fig:fig10}
\end{figure*}

\figref{fig8} show some examples that illustrate the scores of $U_A$, $U_B$, $U_{AB}$
and $U_m$ individually. \figref{fig9} lists the scores of $U$ for all $35$ primitive
cases of parallel plots.

\subsection{Comparison with Human Estimation}
\label{sec:Humans}

We consulted 29 volunteers, including 11 visualization researchers and
18 with statistics, mathematics, humanities and non-academic backgrounds.
We asked them how they would make a comparative judgement about visual complexity.
Through a web-based interface, participants compared pairs of randomly generated
primitive cases. For each pair placed side-by-side, the participants were asked make
a choice among three options: ``\emph{Left is less complex than Right}'', ``\emph{Left
is more complex than Right}'', or ``\emph{Left and Right have similar complexity}''.
Participants compared 50 scatter plot pairs and 50 parallel coordinate pairs in two
trial.

We purposely did not introduce the term \emph{uncertainty} to the participants as the
interpretation of each primitive case can be made uncertain for the given information
as long as there is sufficient time. Instead, we simply consult the participants about
which ``case is more visually confusing than another''. We left the participants to
make their own judgement of the definition of term \emph{confusing}, hence the meaning of
\emph{visual complexity}.

The majority of participants used their intuition to compare pairs of patterns.
\figref{fig9} shows 24 bar charts for each scatter plot case and \figref{fig10}
shows 35 bar charts for each parallel coordinate case. In each bar chart, the $k=0$-bar
indicates the number of times when the observer made the same judgement as the
estimation scheme in \secref{2-Cluster} when comparing this specific case $A$ against
a case $X$ randomly selected from the 35 primitive cases. The $+k$ bars
indicates when the observer over estimated the complexity for our scoring of a case,
while $-k$ bars indicate the observers under estimate of the complexity. The
estimation scheme scores $A$ $k$ points higher than $X$. We consider a $[-0.1,+0.1]$
error in judgement an acceptable threshold for determining consistency of human 
observed measures for our scoring system.

The judgements by different human observers are not consistent. We observe for 
the scatter plots from \figref{fig9} that for most cases the distribution is clustered
around the 0-bar with an error of approximately [-0.5,+0.5]. This is surprising, as
scatter plots are generally considered to be a simpler data representation than parallel
coordinates. The distributions are spread broadly with a few cases showing noticeable over
estimates, (cases 2, 7, 9, 10) or under estimtes (cases 5, 6, 19). This suggests that
although topologically, scatterplots are the simpler representation, observers have
difficulty in judging the relationships between clusters on orthogonal axes. 
In contrast, \figref{fig10} shows tighter clusters more consistent with the parallel 
coordinates complexity scoring. There are more obvious overestimating cases
(7, 9, 11, 16, 23), and a few underestimating cases (5, 30, 32). Overall human
estimations are less dispersed with the parallel coordinates possibly because the
topology is more constrained as clusters are limited in where they appear on
parallel axes.

We examined these cases in detail. Some inconsistency can be explained. For example,
the underestimation in case 1 for both plots is largely because the participants
mistook the two totally overlapped shapes as a single shape. This actually confirms
that the computer score of 1 is correct.  We also made attempts to alter the estimation
scheme. However, we could not find a better scheme, as each attempted change only
resulted in more over- or under-scores in other cases. We believe that this is an
interesting research problem for future work. One possibility is to conduct a large
scale collection of the judgements of human observers. From such empirical data, one
may be able to establish a better estimation scheme, or simply make the mean values
of the human observations as the scores. Such an empirical study is beyond the scope
of this work.

\subsection{Approximating $n$-cluster Complexity}
\label{sec:n-Cluster}
In practice, both scatter plots and parallel coordinates are expected to handle more
than 2 clusters. The extension of Allen's interval algebra from a 2-operand algebra
to an $n$-operand algebra is a non-trivial challenges. We hence address this need by
approximating $n$-cluster complexity by making use of the 2-cluster estimation scheme.
Let $A_1, A_2, \ldots, A_n$ be $n$ clusters. Let $\sigma_2(A_i, A_j)$ be a 2-cluster
score of the pair $A_i$ vs. $A_j$, where $i, j = 1, 2, \ldots, n$. We can approximate
the $n$-cluster score $\sigma_n$ as:
\[  
\sigma_n(A_1, A_2, \ldots, A_n) = \frac{2}{n(n-1)} \sum_{i=1}^{n-1} \sum_{j=i+1}^{n} \sigma_2(A_i, A_j) 
\]
The average is also a quantity $\in [0.0, 1.0]$ as $\sigma_2(A_i, A_j)$.

\section{Conclusions \& Future Work}
\label{sec:conclusions}
This is a theoretical study on visual complexity in the context of cluster visualization.
The central thesis is that it is possible to use Allen's interval algebra to derive a
scheme for estimating visual complexity. As this is an ambitious thesis, this work is
merely the first step to bring mathematics and user experience together. We have
confirmed, both manually and computationally, the primitive cases in 2D Allen's
interval algebra, which is useful for reducing the look-up cases for the estimation
scheme. We have formulated estimation schemes for scatter plots and parallel
coordinates plots. We have collected some human estimations about visual complexity
in relation to these two plots. In comparison with the subjective judgements by humans, our estimation
schemes ar promising.

This research points to a number of interesting and challenging directions for future
studies. These include the need for us to gain further understanding about how humans
estimate visual complexity (e.g., how geometry and topology interfere with each other).
As the sampling space is fairly large (e.g., $35 \times 35$ for parallel coordinates
plots), this would require a large scale empirical study with carefully designed stimuli.
We hope to continue this work, and use both mathematics and empirical studies to create
quantiative metrics for visualization.

\bibliographystyle{eg-alpha}
\bibliography{references}

\newcommand{\etalchar}[1]{$^{#1}$}
\begin{thebibliography}{\uppercase{AdOL04}}

\bibitem[AA04]{Andrienko2004}
\textsc{Andrienko G., Andrienko N.}:
\newblock Parallel coordinates for exploring properties of subsets.
\newblock In \emph{Proc. Coordinated and Multiple Views in Exploratory
  Visualization} (2004), IEEE, pp.~93--104.

\bibitem[AdOL04]{Artero2004}
\textsc{Artero A.~O., de~Oliveira M. C.~F., Levkowitz H.}:
\newblock Uncovering clusters in crowded parallel coordinates visualizations.
\newblock In \emph{Proc. IEEE Information Visualization} (2004), IEEE,
  pp.~81--88.

\bibitem[All83]{allen1983}
\textsc{Allen J.~F.}:
\newblock Maintaining knowledge about temporal intervals.
\newblock \emph{Communications of the ACM} (Novemember 1983), 832--843.

\bibitem[BS06]{Bertini2006}
\textsc{Bertini E., Santucci G.}:
\newblock Visual quality metrics.
\newblock In \emph{Proc. BELIV Workshop} (2006), pp.~1--5.

\bibitem[CDD06]{carr2006}
\textsc{Carr H., Duffy B., Denby B.}:
\newblock On histograms and isosurface statistics.
\newblock \emph{IEEE Transactions on Visualization and Computer Graphics 12}, 5
  (2006), 1259--1266.

\bibitem[DCK12]{dasgupta:2012}
\textsc{Dasgupta A., Chen M., Kosara R.}:
\newblock Conceptualizing visual uncertainty in parallel coordinates.
\newblock \emph{Computer Graphics Forum 31}, 3pt2 (june 2012), 1015--1024.

\bibitem[DCT12]{duffy2012}
\textsc{Duffy B., Carr H., Torsten M.}:
\newblock Integrating isosurface statistics and histograms.
\newblock \emph{IEEE Transactions on Visualization and Computer Graphics 14}
  (2012).

\bibitem[DK10]{Dasgupta:InfoVis:2010}
\textsc{Dasgupta A., Kosara R.}:
\newblock Pargnostics: Screen-space metrics for parallel coordinates.
\newblock \emph{IEEE Transactions on Visualization and Computer Graphics 16}, 6
  (2010), 1017--26.

\bibitem[DK11]{Dasgupta:InfoVis:2011}
\textsc{Dasgupta A., Kosara R.}:
\newblock Adaptive privacy-preservation using parallel coordinates.
\newblock \emph{IEEE Transactions on Visualization and Computer Graphics 17},
  12 (2011), 2241--2248.

\bibitem[ED06]{Ellis2006}
\textsc{Ellis G., Dix A.}:
\newblock {Enabling automatic clutter reduction in parallel coordinate plots}.
\newblock \emph{IEEE Transactions on Visualization and Computer Graphics 12}, 5
  (2006), 717--724.

\bibitem[ED07]{ellis}
\textsc{Ellis G., Dix A.}:
\newblock A taxonomy of clutter reduction for information visualisation.
\newblock \emph{IEEE Transactions on Visualization and Computer Graphics, 13},
  6 (2007), 1216--1223.

\bibitem[FWR99]{Fua:Vis:1999}
\textsc{Fua Y.-H., Ward M.~O., Rundensteiner E.~A.}:
\newblock Hierarchical parallel coordinates for exploration of large datasets.
\newblock In \emph{Proc. IEEE Visualization} (1999), IEEE CS Press, pp.~43--50.

\bibitem[HMS09]{Harper:2009:TDV:}
\textsc{Harper S., Michailidou E., Stevens R.}:
\newblock Toward a definition of visual complexity as an implicit measure of
  cognitive load.
\newblock \emph{ACM Transactions on Applied Perception 6}, 2 (2009),
  10:1--10:18.

\bibitem[ID90]{Inselberg1990}
\textsc{Inselberg A., Dimsdale B.}:
\newblock Parallel coordinates: A tool for visualizing multi-dimensional
  geometry.
\newblock In \emph{Proc. IEEE Visualization} (1990), IEEE CS Press,
  pp.~361--378.

\bibitem[JLJC05]{Johansson}
\textsc{Johansson J., Ljung P., Jern M., Cooper M.}:
\newblock {Revealing structure within clustered parallel coordinates displays}.
\newblock In \emph{Proc. IEEE Symposium on Information Visualization} (2005),
  pp.~125--132.

\bibitem[KW10]{khoury2010}
\textsc{Khoury M., Wenger R.}:
\newblock On the fractal dimension of isosurfaces.
\newblock \emph{IEEE Transactions on Visualization and Computer Graphics 16}, 6
  (Nov.-Dec. 2010), 1198--1205.

\bibitem[LC87]{lorensen1987}
\textsc{Lorensen W.~E., Cline H.~E.}:
\newblock Marching cubes: a high resolution 3{D} surface construction
  algorithm.
\newblock In \emph{Proc. ACM SIGGRAPH} (New York, NY, USA, 1987), pp.~163--169.

\bibitem[NH06]{Novotny2006}
\textsc{Novotny M., Hauser H.}:
\newblock Outlier-preserving focus+context visualization in parallel
  coordinates.
\newblock \emph{IEEE Transactions on Visualization and Computer Graphics 12}, 5
  (2006), 893--900.

\bibitem[PWR04]{Peng2004}
\textsc{Peng W., Ward M., Rundensteiner E.}:
\newblock {Clutter reduction in multi-dimensional data visualization using
  dimension reordering}.
\newblock In \emph{Proc. IEEE Information Visualization} (2004), IEEE CS Press,
  pp.~89--96.

\bibitem[RLN07]{rosenholtz2007}
\textsc{Rosenholtz R., Li Y., Nakano L.}:
\newblock Measuring visual clutter.
\newblock \emph{Journal of Vision 7}, 2 (2007).

\bibitem[SBS{\c{C}}10]{schnur2010comparison}
\textsc{Schnur S., Bekta{\c{s}} K., Salahi M., {\c{C}}{\"o}ltekin A.}:
\newblock A comparison of measured and perceived visual complexity for dynamic
  web maps.
\newblock In \emph{Proc. 6th International Conference on Geographic Information
  Science} (2010).

\bibitem[SSD{\etalchar{*}}08]{scheidegger2008}
\textsc{Scheidegger C.~E., Schreiner J.~M., Duffy B., Carr H., Silva C.~T.}:
\newblock Revisiting histograms and isosurface statistics.
\newblock \emph{IEEE Transactions on Visualization and Computer Graphics 14}, 6
  (2008), 1659--1666.

\bibitem[ZYQ{\etalchar{*}}08]{Zhou2008}
\textsc{Zhou H., Yuan X., Qu H., Cui W., Chen B.}:
\newblock {Visual clustering in parallel coordinates}.
\newblock \emph{Computer Graphics Forum 27}, 3 (2008), 1047--1054.

\end{thebibliography}

\end{document}